\documentclass{article}

\usepackage[final]{neurips_2022}

\usepackage[utf8]{inputenc} 
\usepackage[T1]{fontenc}    
\usepackage{hyperref}       
\usepackage{url}            
\usepackage{booktabs}       
\usepackage{amsfonts}       
\usepackage{nicefrac}       
\usepackage{microtype}      
\usepackage{xcolor}         
\usepackage{graphicx}
\usepackage{subcaption}
\usepackage{float}
\usepackage{amsmath}
\usepackage{caption}
\title{Exploring Grokking: Experimental and Mechanistic Investigations}

\author{
  QiYe Hu\\
  \texttt{2000010846}
  \And
  Hao Zhou\\
  \texttt{2301110061}
  \And
  RuoXi Yu \\
  \texttt{2301111452}
}

\begin{document}

\maketitle

\begin{abstract}

The phenomenon of \emph{grokking} in over-parameterized neural networks has garnered significant interest. It involves the neural network initially memorizing the training set with zero training error and near-random test error. Subsequent prolonged training leads to a sharp transition from no generalization to perfect generalization. Our study comprises extensive experiments and an exploration of the research behind the mechanism of grokking. Through experiments, we gained insights into its behavior concerning the training data fraction, the model, and the optimization. The mechanism of grokking has been a subject of various viewpoints proposed by researchers, and we introduce some of these perspectives.
\end{abstract}

\section{Introduction}
The mysterious behavior of over-parameterized neural networks has become a fascinating topic. Despite their tendency to overfit the training set, these networks often exhibit remarkable consistency in their training and testing performance when using popular gradient-based optimizers. It is generally believed that this phenomenon is due to the inherent properties of network architectures and training methods that have the inherent ability to autonomously instill regularization effects.

Recently, Power et al. \cite{power2022grokking} uncovered a perplexing generalization phenomenon known as \emph{grokking}. In the context of training a neural network to learn modular arithmetic operations, grokking involves the initial "memorization" of the training set with zero training error and near-random test error. Prolonged training then induces a sharp transition from no generalization to perfect generalization.

In addition to modular arithmetic, the phenomenon of grokking has been observed in various research domains. Grokking has been reported in learning group operations, sparse parity \cite{chughtai2023toy}, learning the greatest common divisor, image classification \cite{charton2023transformers, barak2023hidden, bhattamishra2023simplicity, liu2022omnigrok, radhakrishnan2023mechanism}.


Different perspectives on the mechanism of grokking have been suggested. These include the slingshot mechanism, which involves cyclic phase transitions Thilak et al., 2022\cite{thilak2022slingshot}; the concept of slow formulation of good representations Liu et al., 2022\cite{liu2022understanding}; the influence of the scale of initialization Liu et al., 2023\cite{liu2022omnigrok}; and the significance of the simplicity of the generalizable solution, as discussed by Nanda et al. 2023\cite{nanda2023progress} and Varma et al. 2023\cite{varma2023explaining}.

The organization of the article content is as follows. Section \ref{section:setting} defines our task of representing a binary operation using a neural network and outlines our method of encoding characters. Section \ref{section:modelopt} introduces the models, optimizers, and hyperparameters used in our experiments. The results and analysis of our experiments are presented in Section \ref{section:exp}. Finally, we delve into the mechanisms of grokking in Section \ref{section:grok}.

\section{Problem Setting}
\label{section:setting}
\subsection{Modular Addition and Training Data Fraction}
In this research project, we investigate the modular addition operation within the context of binary operations. The modular addition operation is defined as follows:
$$(x, y)\to (x + y)\% \,p \ \text{  for}\  x, y \in  \mathbb{Z}_p.$$
The objective is to develop a neural network that can predict the probabilities associated with modular addition results. The network outputs a probability distribution, and the outcome is determined by selecting the highest probability.

The total number of possible input pairs for this study is constrained to $p^2$. To construct the training dataset, a fraction $\alpha$ of these pairs is randomly selected, while the remaining pairs form the validation set. The training data fraction, denoted as $\alpha$, is defined as:
$$\alpha=\frac{\text{size of training data}}{\text{size of total data}}.$$

When the value of the training data fraction $\alpha$ is small, it is possible to achieve perfect accuracy on the training set; however, this may result in significantly lower prediction accuracy on the validation set. In other words, the model may struggle to generalize beyond the training data, posing challenges in addressing the binary operation problem. As $\alpha$ increases, the difficulty of achieving perfect accuracy on the training set also increases, while the difficulty of solving the problem or achieving complete generalization decreases.

\subsection{Encode}\label{implement}
This project involves exploring two methodologies for translating mathematical operations into classification or sequence continuation problems suitable for neural network processing. These approaches draw inspiration from the methods outlined in the article by Power et al. \cite{power2022grokking} and utilize the official code repository at \url{https://github.com/openai/grok} as well as the implementation provided in the Colab notebook available at \url{https://colab.research.google.com/github/neelnanda-io/TransformerLens/blob/main/demos/Grokking_Demo.ipynb}.

In the article \cite{power2022grokking}, various binary modular operations and binary operations on the permutation group $S_5$ are considered. The methodology involves creating a comprehensive dictionary that encompasses all relevant numerical entities, including elements from $\mathbb{Z}_p$ and $S_5$. The operations, which include modulo-based operations such as addition, subtraction, and multiplication, as well as connecting characters like space and $=$, are also included in this dictionary. Each element in the dictionary is assigned an index, and these indices are used as the actual input and output values.

For instance, if the indices assigned to $1,2,3,+,=$ are $130,131,132,3,10$ respectively, then to compute $(1+2)\%p = 3$, the input would be $[130,3,131,10]$, and the output would be $132$. The adoption of this indexing approach, which does not directly involve the representation of "integers", introduces additional complexity to the task of learning the underlying operations.

The implementation offered in the Colab notebook notably streamlines the task. For modular addition, it accepts three input values, encompassing the integers $x$, $y$ and 
$p$, where $0 \leq x, y < p$. The mathematical operation is framed as a classification task, presenting probabilities for each of the 
$p$ potential outcomes.

\section{Model and Optimization}
\label{section:modelopt}
In this project, three distinct models will be employed: Transformer, LSTM, and MLP. 
In this section, we provide brief overviews of the models. and introduce the model parameters and optimization method parameters used in the subsequent experiments.

\paragraph{Transformer}: The Transformer model, introduced by Vaswani et al. in the seminal paper \cite{vaswani2017attention}, is renowned for its efficiency in processing sequential data. Key components include the embedding layer, attention mechanism, multi-head attention, and positional encoding. Its parallelized processing and self-attention mechanism make it particularly effective in capturing complex relationships within sequences

For the experiments conducted in \cite{power2022grokking}, a standard decoder-only Transformer with causal attention masking was employed. The model featured 2 layers, a width of 128, and 4 attention heads, totaling approximately $4 \cdot 10^5$ non-embedding parameters. Optimization was performed using the AdamW optimizer with a learning rate of $10^{-3}$, weight decay set to 1, $\beta_1 = 0.9, \beta_2 = 0.98$, linear learning rate warm-up executed over the initial 10 updates, a minibatch size of 512, and an optimization budget of $10^5$ gradient updates.

For the simplified Transformer model, the configuration included a one-layer MLP with a hidden size of 512 for the feed-forward networks. The multi-head attention utilized 4 attention heads with $H=32$ and $d_{model}=128$, and the activation function was set to ReLU.
The optimizer's parameters were consistent with the aforementioned model, with one notable decision: the model was trained using full-batch training instead of stochastic gradient descent. This choice aimed to achieve smoother training dynamics and reduce the occurrence of slingshots during optimization.

\paragraph{LSTM (Long Short-Term Memory)}: LSTM, a variant of recurrent neural networks, is designed to overcome challenges like vanishing gradients in sequential data processing. It incorporates LSTM cells with forget, input, and output gates, allowe dependencies over extended sequences. Its sequential processing capability makes it suitable for various tasks, such as natural language processing and time-series prediction.

We employed an LSTM with a hidden size of 20. Optimization was performed using the Adam optimizer with a learning rate of $10^{-2}$.

\paragraph{MLP (Multi-Layer Perceptron)}: MLP is a type of neural network without memory. It learns complex patterns in data through nonlinear transformations between fully connected layers. Each layer applies an activation function to the weighted sum of its inputs. MLP is good for non-sequence tasks like image classification. Unlike LSTM, it cannot directly handle time series data without feature transformations.

We employed an MLP with a hidden size of 512, featuring 2 layers and utilizing the ReLU activation function. Optimization was performed using the Adam optimizer with a learning rate of $10^{-3}$.

We delve further into optimization details in Subsection \ref{subsection:opt}.

\section{Experiments}
\label{section:exp}
This section presents and analyzes our experiments with modular addition in $p=97$. The three parts include exploration of the model's generalization performance when varying the training data fraction, examination of the behavior in the context of different models and investigation into the impact of utilizing different optimization methods during training.

\subsection{Generalization Across Varied Training Data Fractions}
As outlined in Section \ref{implement}, two implementation methods were presented. In this section, we showcase the generalization performance when altering the training data fraction using both methods. Additionally, we highlight the observed grokking phenomena.

\begin{figure}[!ht]
    \begin{subfigure}{0.32\linewidth}
        \centering
        \includegraphics[width=4cm,height=3cm]{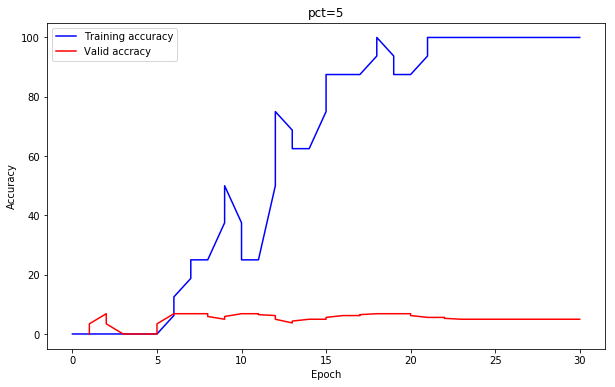}
        \caption{Accuracy at 5\% training data.}
    \end{subfigure}
    \begin{subfigure}{0.32\linewidth}
        \centering
        \includegraphics[width=4cm,height=3cm]{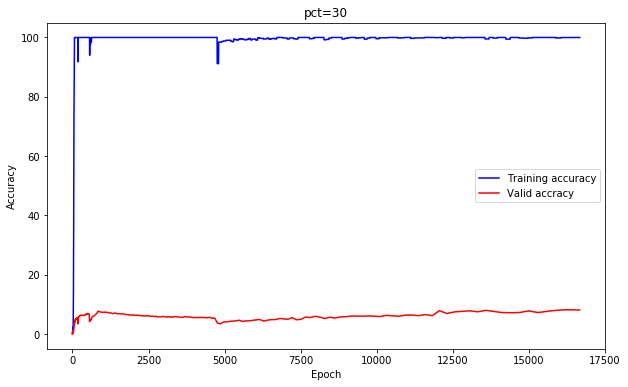}
        \caption{Accuracy at 30\% training data.}
    \end{subfigure}
    \begin{subfigure}{0.32\linewidth}
        \centering
        \includegraphics[width=4cm,height=3cm]{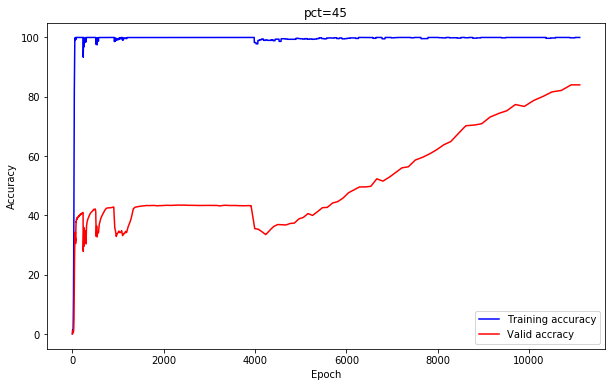}
        \caption{Accuracy at 45\% training data.}
    \end{subfigure}
    
    \medskip
    
    \begin{subfigure}{0.32\linewidth}
        \centering
        \includegraphics[width=4cm,height=3cm]{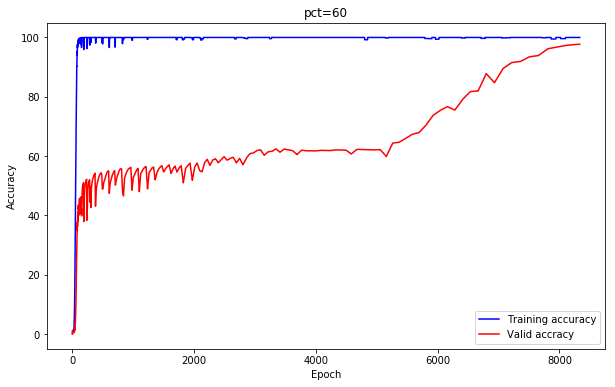}
        \caption{Accuracy at 60\% training data.}
    \end{subfigure}
    \begin{subfigure}{0.32\linewidth}
        \centering
        \includegraphics[width=4cm,height=3cm]{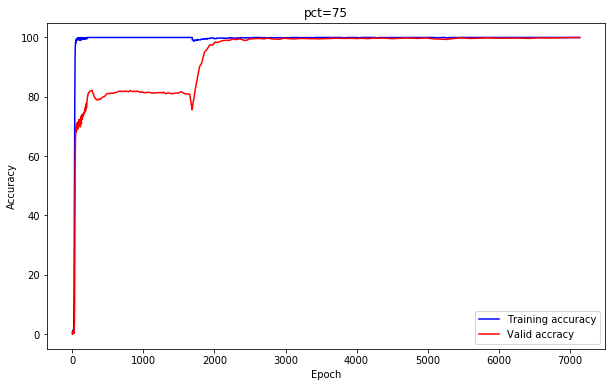}
        \caption{Accuracy at 75\% training data.}
    \end{subfigure}
    \begin{subfigure}{0.32\linewidth}
        \centering
        \includegraphics[width=4cm,height=3cm]{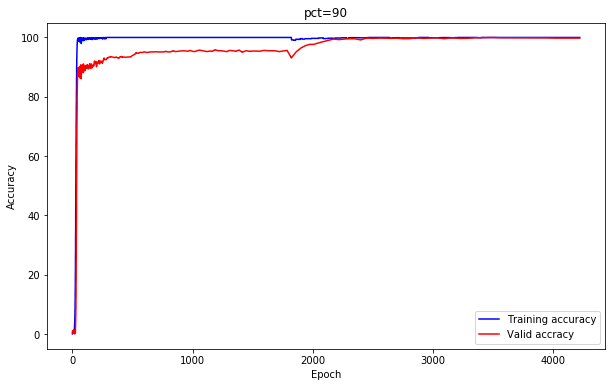}
        \caption{Accuracy at 90\% training data.}
    \end{subfigure}    
    \caption{Comparison of accuracy at different training data fractions with mod 97. The blue lines for training and the red lines for validation.}
    \label{fig:acc_step_1}
\end{figure}

We execute the source code to obtain data for Figure \ref{fig:acc_step_1}. The figure illustrates when $\alpha$ is small, the task may be too hard for the Transformer to handle, and validation accuracy starts to grow until $\alpha$ reaches around $40\%$. This phenomena is mentioned in Figure2 of the original article \cite{power2022grokking}. Also, when $\alpha$ takes $45\%$ and $60\%$, the gap between the two curves is obvious and it takes around 6000 epoch for generalization. This means we have arrived at the same conclusion as the original article \cite{power2022grokking}, that the grokking phenomenon is most pronounced when alpha is around $50\%$. When $\alpha$ takes $75\%$ and $90\%$, valid accuracy grows almost as fast as training accuracy and we think this is a certain outcome of a finite binary operation table and an excessive amount of known information.

\begin{figure}[!ht]
    \begin{subfigure}{0.32\linewidth}
        \centering
        \includegraphics[width=4cm,height=3cm]{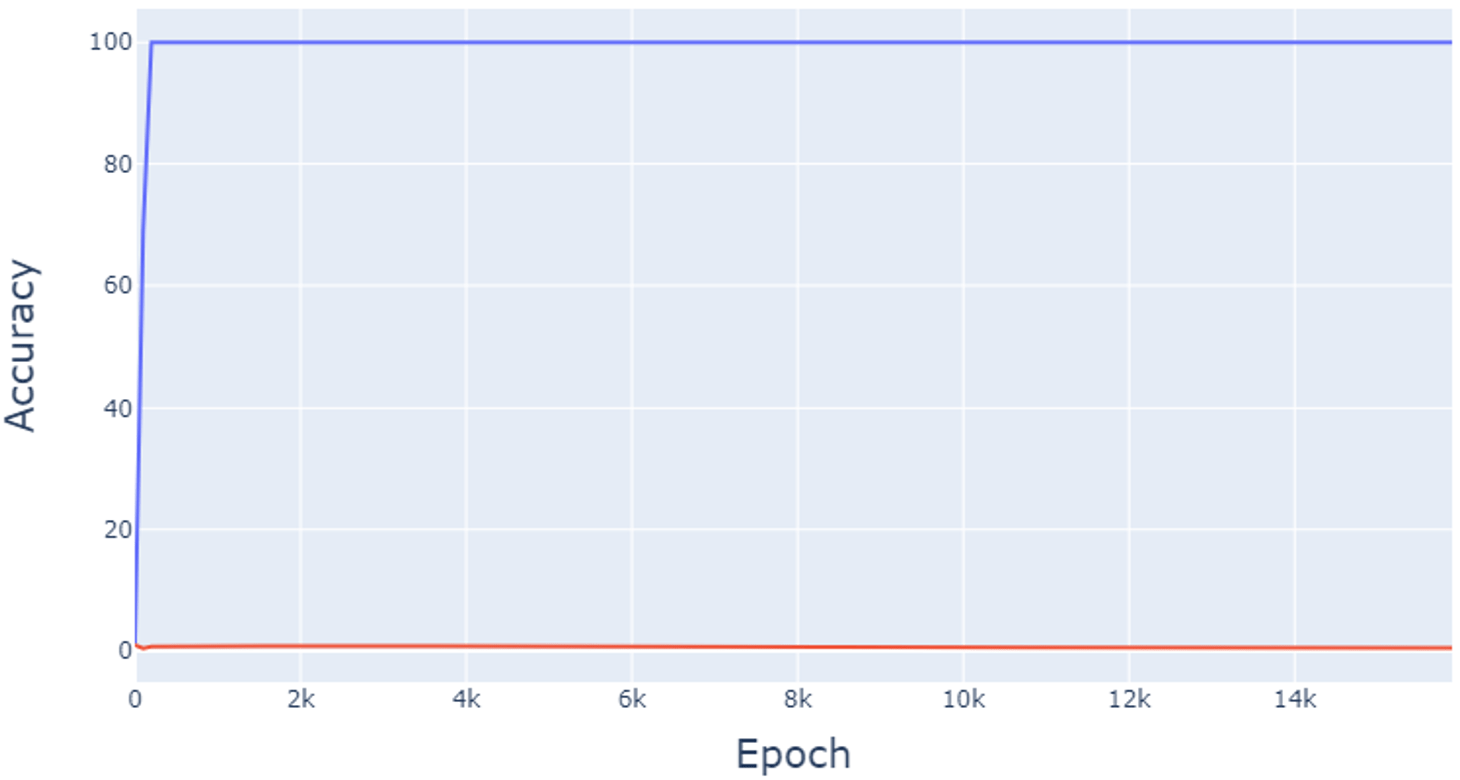}
        \caption{Accuracy at 15\% training data.}
    \end{subfigure}
    \begin{subfigure}{0.32\linewidth}
        \centering
        \includegraphics[width=4cm,height=3cm]{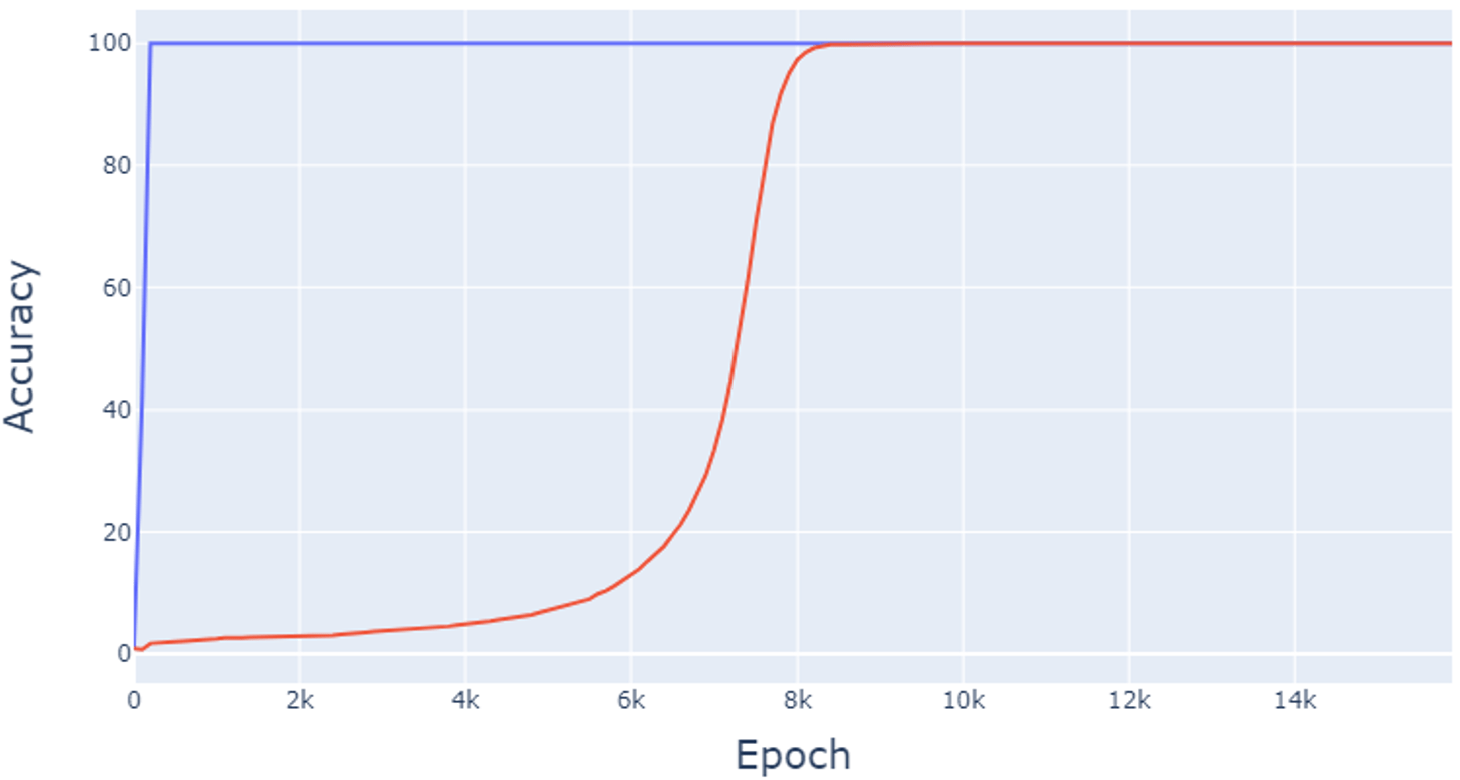}
        \caption{Accuracy at 30\% training data.}
    \end{subfigure}
    \begin{subfigure}{0.32\linewidth}
        \centering
        \includegraphics[width=4cm,height=3cm]{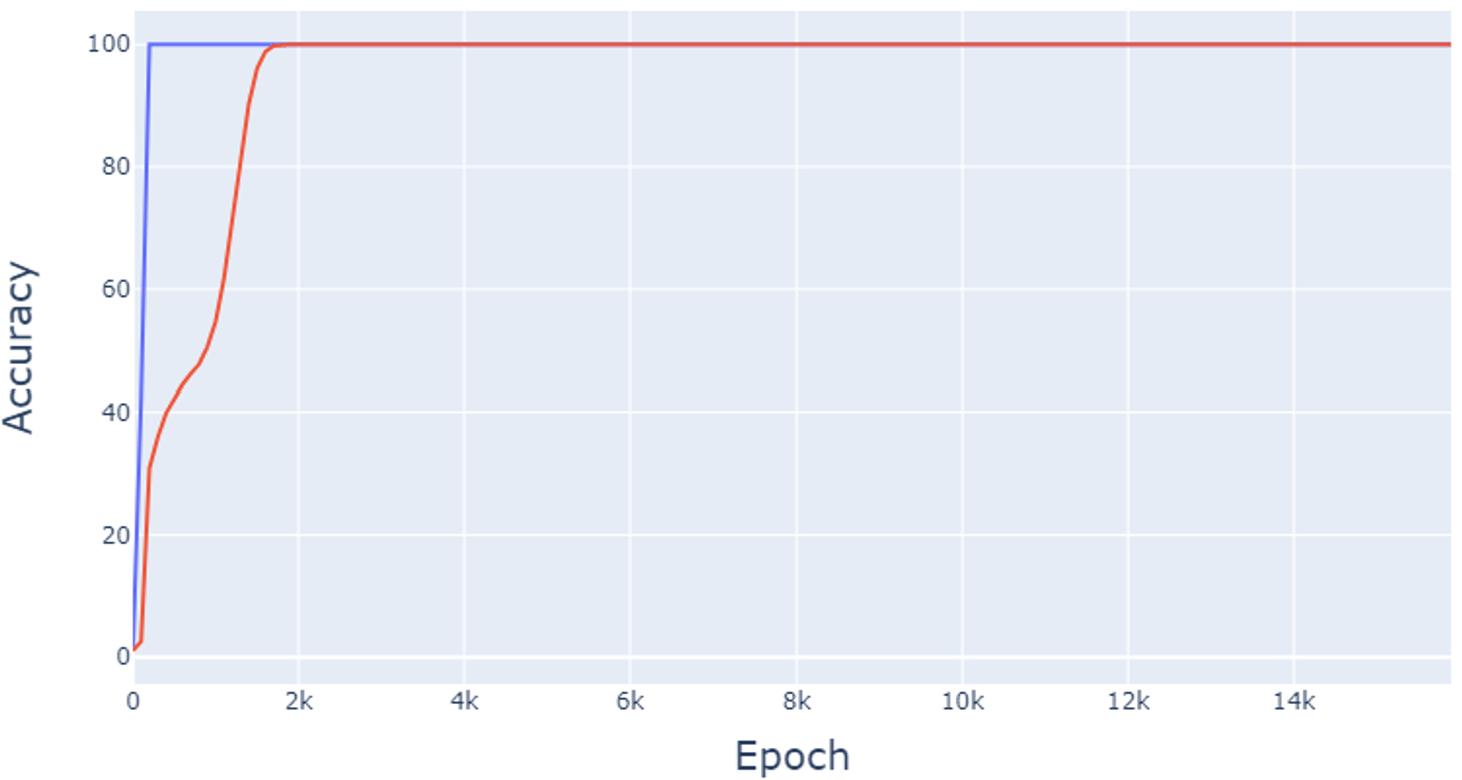}
        \caption{Accuracy at 45\% training data.}
    \end{subfigure}
    \caption{Comparison of accuracy at different training data fractions. The blue lines for training and the red lines for validation.}
    \label{fig:acc_step_trans2}
\end{figure}

Utilizing a straightforward encoding method alongside a Transformer model, Figure \ref{fig:acc_step_trans2} illustrates three typical generalization scenarios in the modular addition problem. For $\alpha=15\%$, we trained for 25k epochs, where the validation data loss showed no improvement, and the accuracy remained close to zero, indicating a failure in generalization. With $\alpha=30\%$, the training data reached 100\% accuracy after around 200 steps, but the validation accuracy surged around 6k steps, reaching 100\% at 9.1k steps, signifying successful and complete generalization. At $\alpha=45\%$, the overall generalization behavior resembled typical cases, with both train and validation curves exhibiting similar trends. However, an observable delay in the validation curve's sharp increase was also noted as was the previous one. 

Additional details regarding the losses and steps until generalization at various training data fractions are presented in Figure \ref{fig:loss_step_trans2} and Figure \ref{fig:step_frac_trans2}.

When comparing different encoding methods, both exhibit the grokking phenomenon. However, the official problem is more challenging, requiring a larger training data fraction for successful generalization.

\subsection{Model Variations}
\label{subsection:exp model}
In this subsection, we employ the straightforward encoding method. We present the accuracy of the MLP and LSTM models at different training data fractions and compare them with the Transformer model.

\begin{figure}[!ht]
    \begin{subfigure}{0.32\linewidth}
        \centering
        \includegraphics[width=4cm,height=3cm]{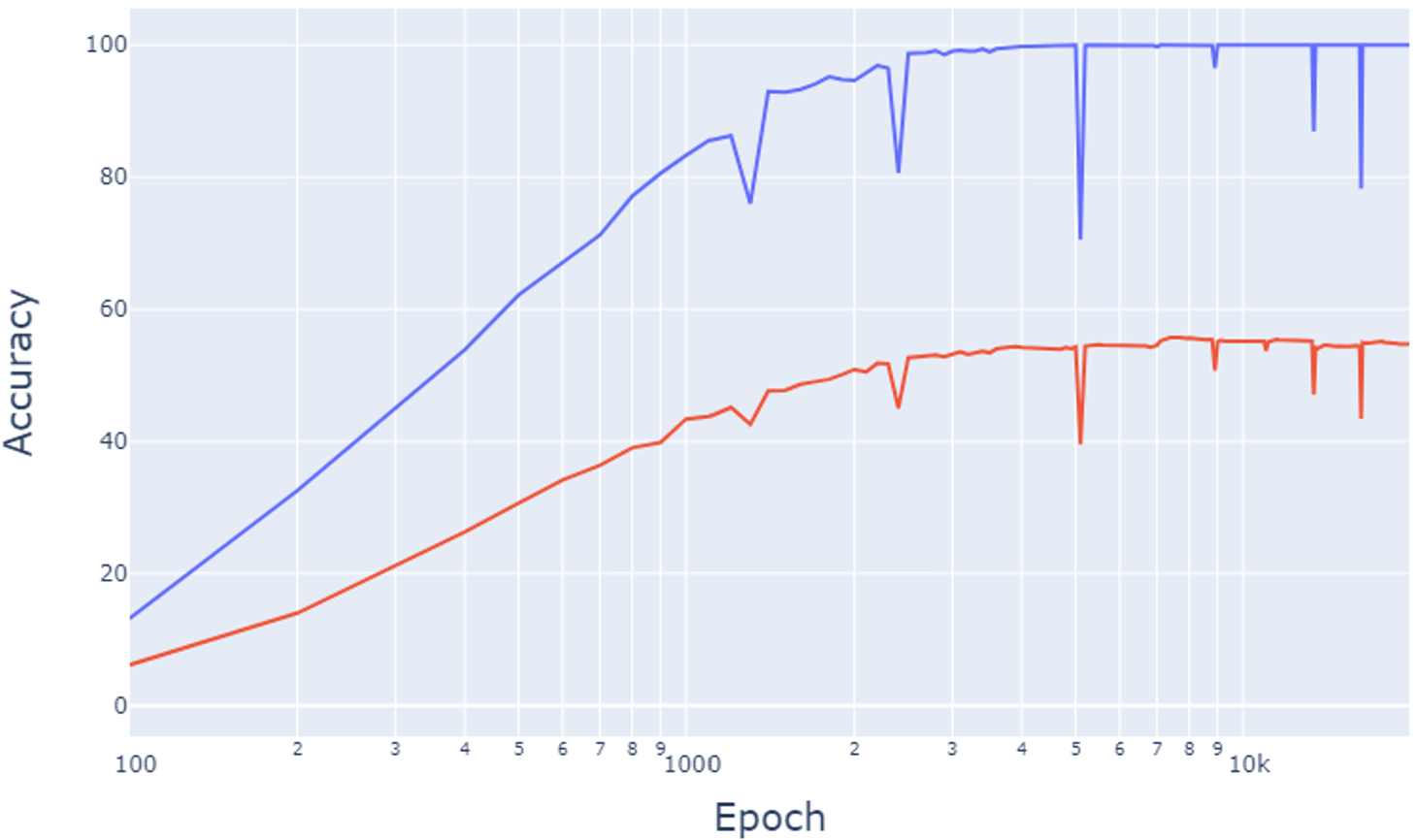}
        \caption{Accuracy at 15\% training data.}
    \end{subfigure}
    \begin{subfigure}{0.32\linewidth}
        \centering
        \includegraphics[width=4cm,height=3cm]{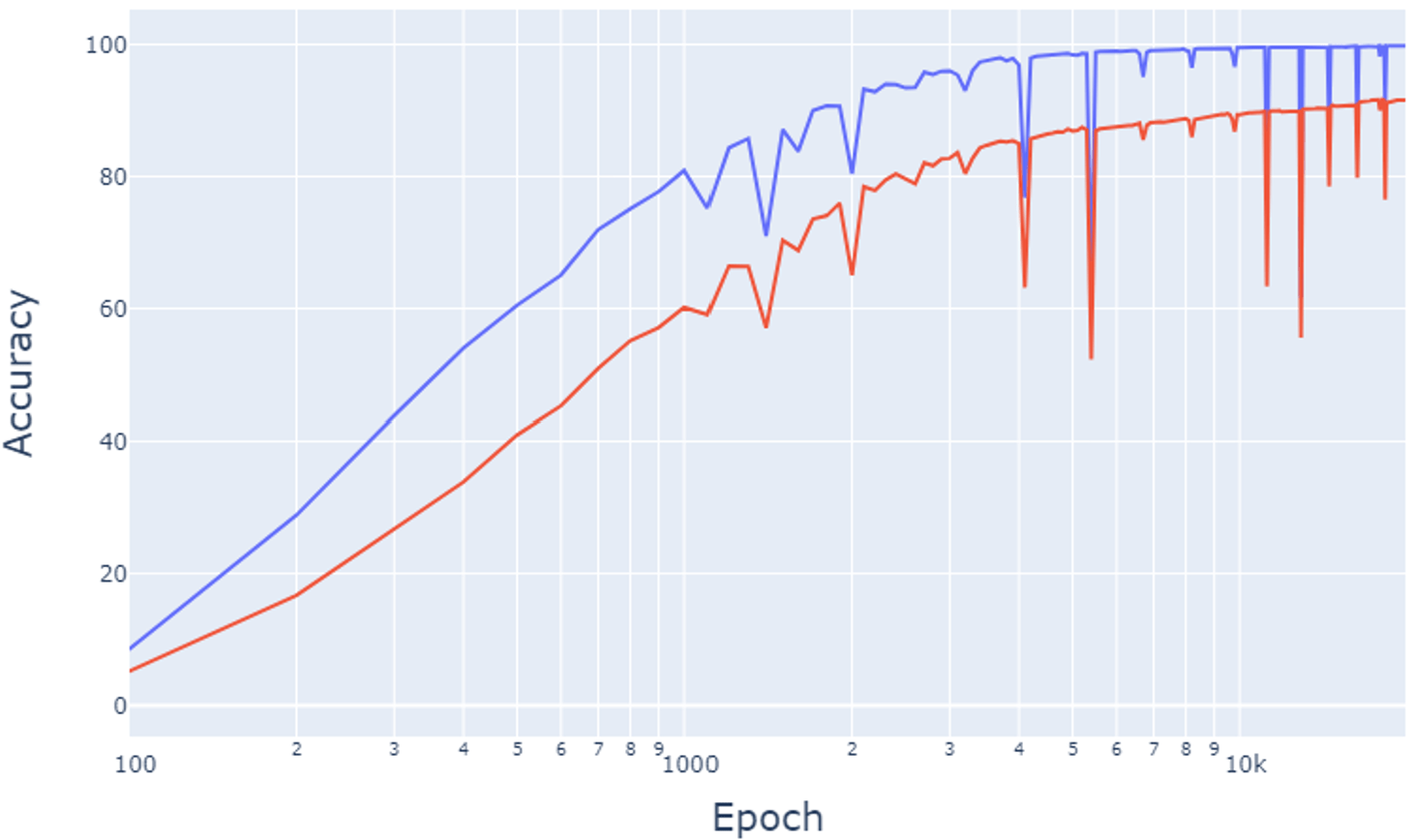}
        \caption{Accuracy at 30\% training data.}
    \end{subfigure}
    \begin{subfigure}{0.32\linewidth}
        \centering
        \includegraphics[width=4cm,height=3cm]{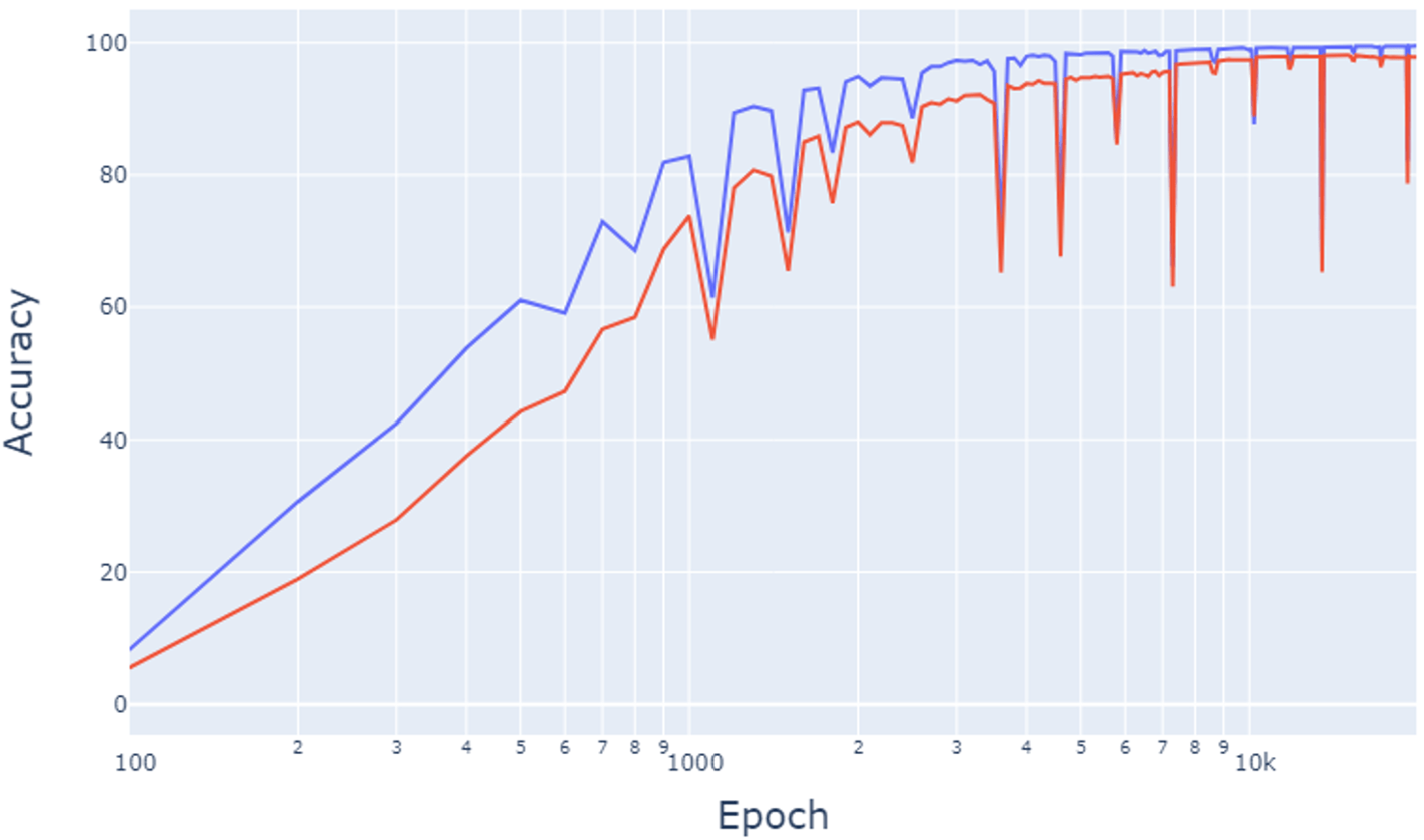}
        \caption{Accuracy at 45\% training data.}
    \end{subfigure}
    \caption{Comparison of accuracy at different training data fractions with MLP model. The blue lines for training and the red lines for validation.}
    \label{fig:acc_step_mlp}
\end{figure}

For the MLP model, we maintained parameters identical to the feed-forward network of the Transformer. The optimization methods were also similar. However, even after 20,000 steps, the training accuracy did not reach 100\%, highlighting the performance of the Transformer in this specific problem.

The three subplots in Figure \ref{fig:acc_step_mlp} depict similar behavior, where training and validation accuracy rise concurrently. When the training accuracy approaches 100\%, the validation accuracy also approaches its maximum achievable value in this experiment. It is evident that the MLP model does not exhibit the grokking phenomenon in this context.


\begin{figure}[!ht]
    \begin{subfigure}{0.32\linewidth}
        \centering
        \includegraphics[width=4cm,height=3cm]{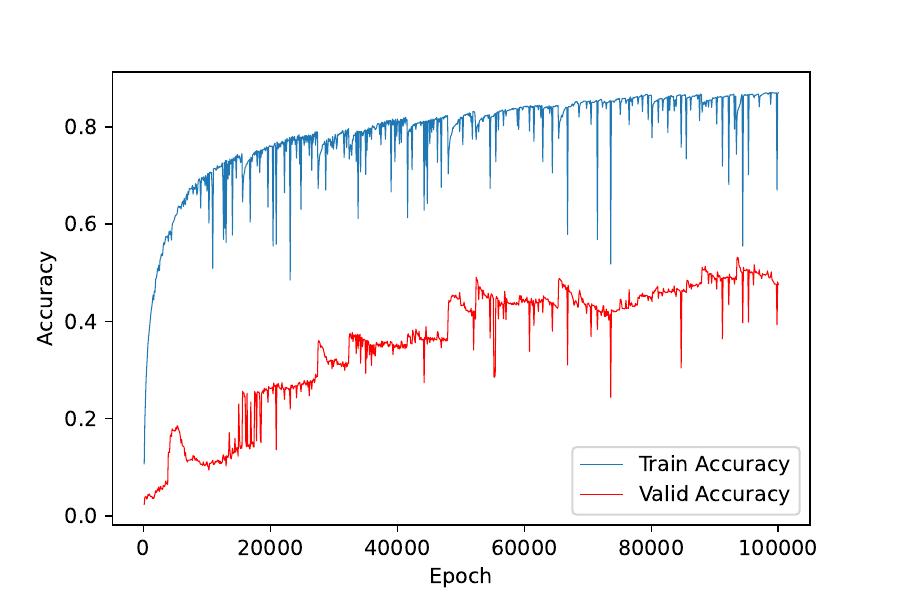}
        \caption{Accuracy at 30\% training data.}
    \end{subfigure}
    \begin{subfigure}{0.32\linewidth}
        \centering
        \includegraphics[width=4cm,height=3cm]{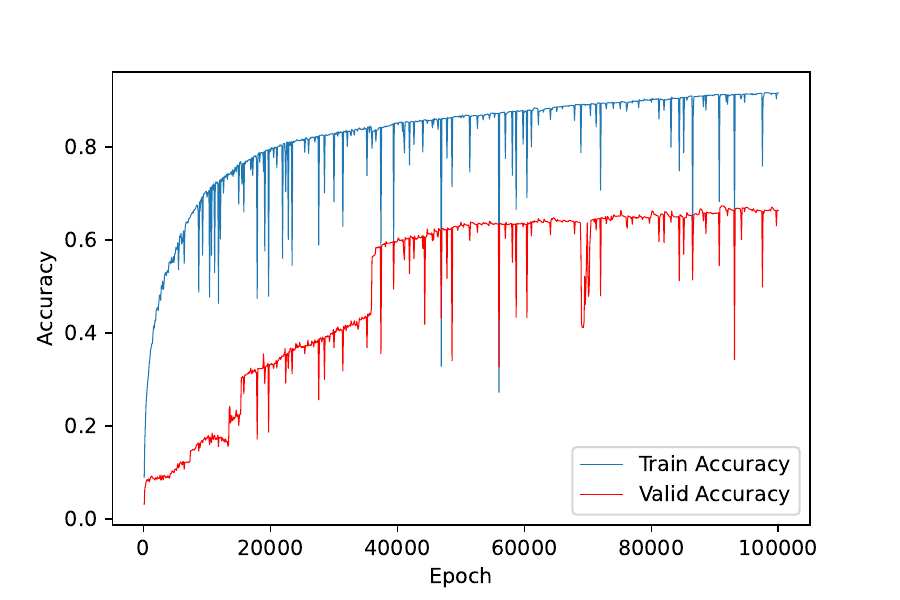}
        \caption{Accuracy at 45\% training data.}
    \end{subfigure}
    \begin{subfigure}{0.32\linewidth}
        \centering
        \includegraphics[width=4cm,height=3cm]{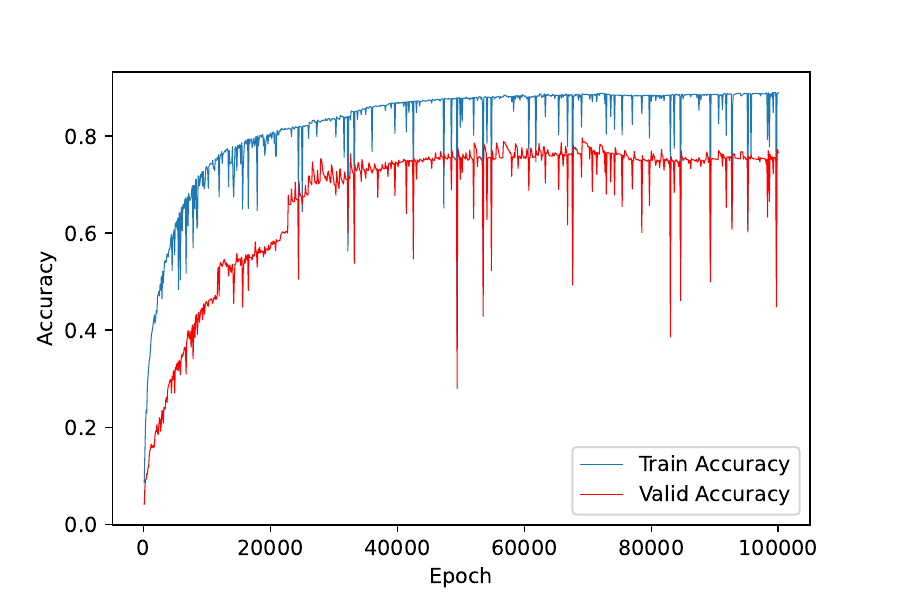}
        \caption{Accuracy at 60\% training data.}
    \end{subfigure}
    \caption{Comparison of accuracy at different training data fractions with LSTM model. The blue lines for training and the red lines for validation.}
    \label{fig:acc_step_LSTM}
\end{figure}

We implement the LSTM model ourselves using the direct encoding method and plot the three charts as shown above. In this experiment, $p$ still takes $97$ and we set the number of hidden layers to $20$ to make both training accuracy and valid accuracy to reach a relatively high level after $100000$ epoch. Since the grokking phenomena is highly related to the percentage of training data, we tried $30\%$, $45\%$ and $60\%$ respectively, but we still cannot get the grokking phenomena.

In summary, we observed the traditional generalization phenomenon in both MLP and LSTM models, which differs from the behavior exhibited by the Transformer model.
However, recent work \cite{liu2022omnigrok} indicates that training with large initialization and small weight decay can induce grokking on various tasks. The Transformer model is not obligatory; the grokking phenomenon can be achieved with clever techniques even on MLPs. We will delve into this discussion in Section \ref{section:grok}.

\subsection{Optimizers and Regularization}
\label{subsection:opt}
Weight decay is a regularization method by adding regular derivative terms in the parameter update process. In the case of standard SGD, L2 regularization and Weight Decay can be regarded as the same by transforming the attenuation coefficient. However, in the adaptive learning rate algorithm such as Adam, the two are not equivalent. The correct way to introduce Weight Decay in Adam is called AdamW.

\begin{figure}[!ht]

    \begin{subfigure}{0.24\linewidth}
        \centering
     \includegraphics[width=2.7cm,height=2cm]{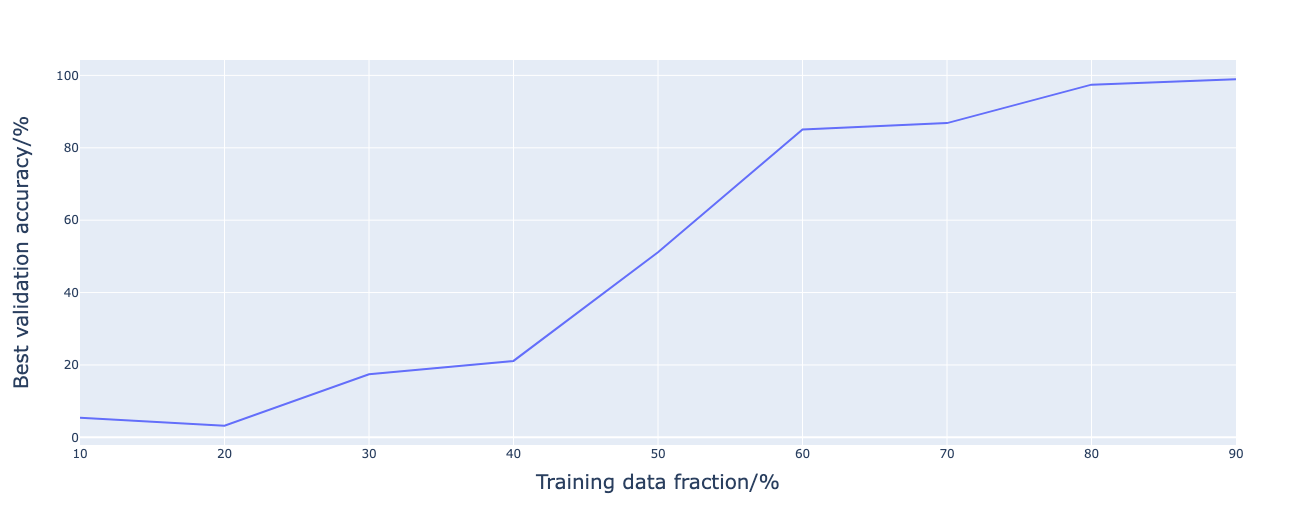}
        \caption{\scriptsize Adam, 3x baseline LR}
    \end{subfigure}
        \begin{subfigure}{0.24\linewidth}
        \centering
    \includegraphics[width=2.7cm,height=2cm]{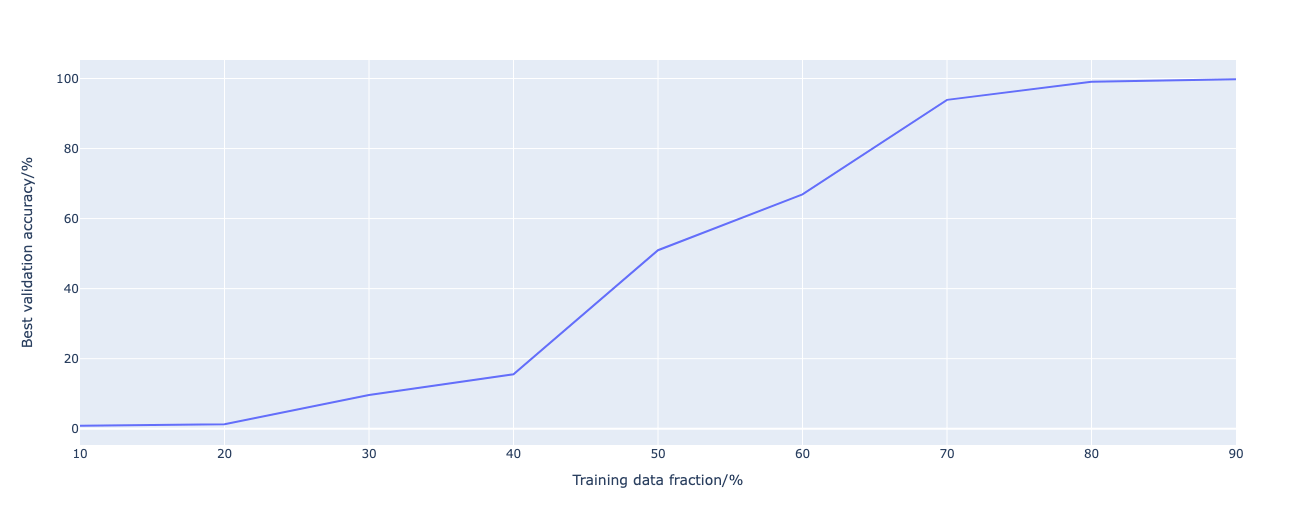}
        \caption{\scriptsize Full Batch Adam, baseline LR}
    \end{subfigure}
    \begin{subfigure}{0.24\linewidth}
        \centering
    \includegraphics[width=2.7cm,height=2cm]{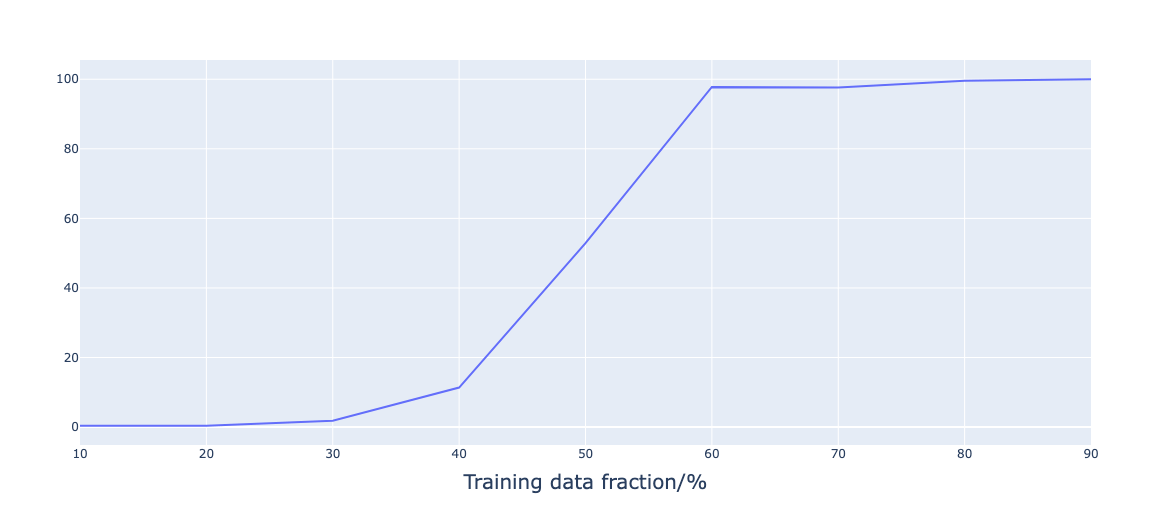}
        \caption{\scriptsize Adam, 0.3x baseline LR}
    \end{subfigure}
    \begin{subfigure}{0.24\linewidth}
        \centering
    \includegraphics[width=2.7cm,height=2cm]{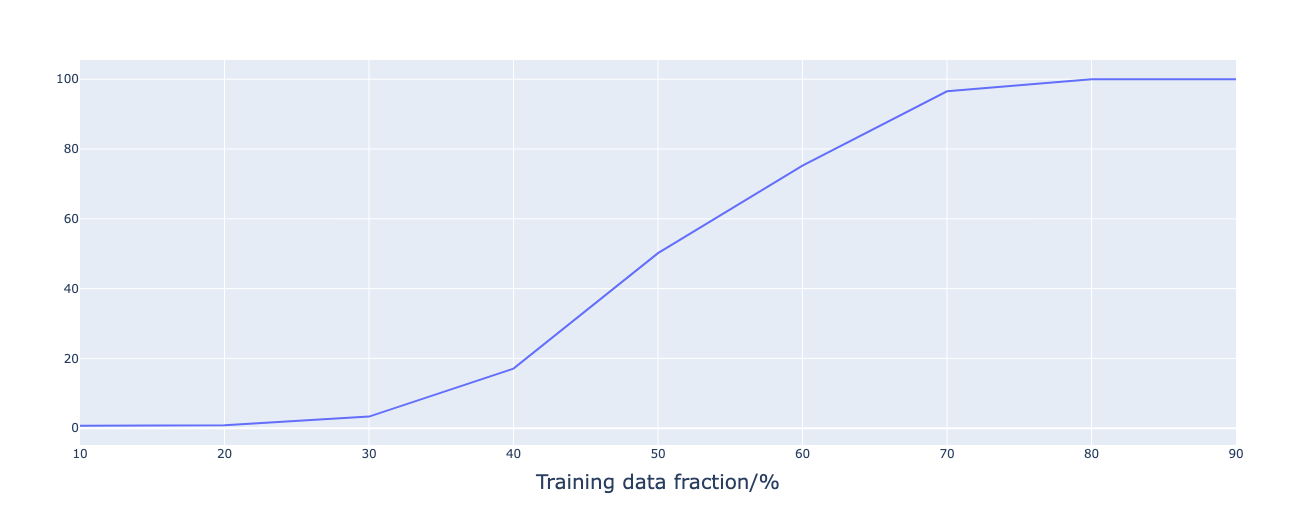}
        \caption{\scriptsize Adam, update direction noise}
    \end{subfigure}
    \medskip
    
    \begin{subfigure}{0.24\linewidth}
        \centering
\includegraphics[width=2.7cm,height=2cm]{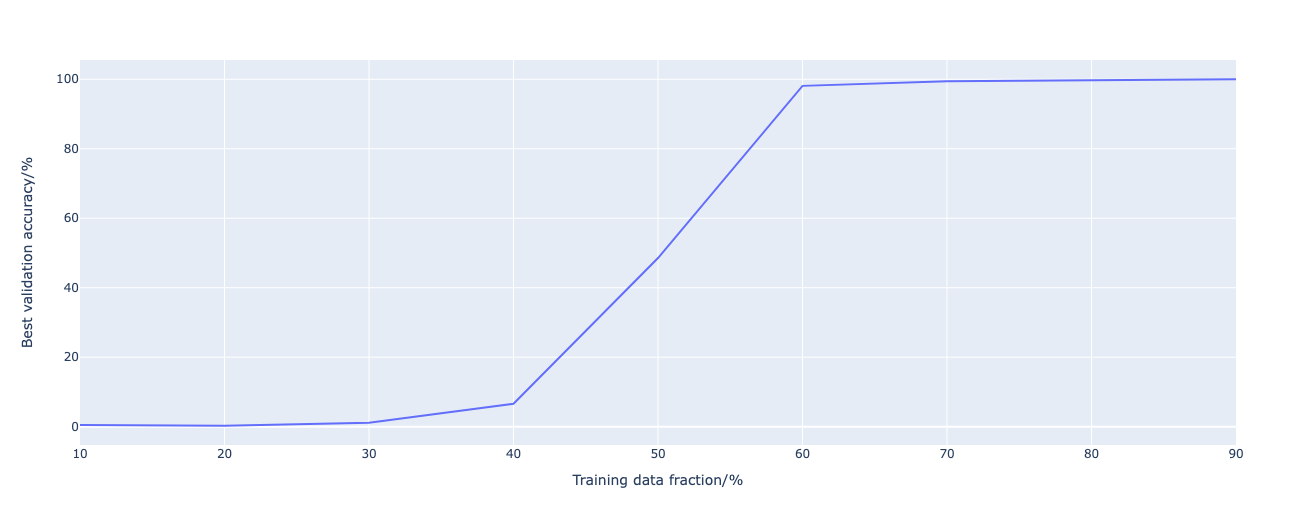}
        \caption{\scriptsize Adam, Gaussian weight noise}
    \end{subfigure}
    \begin{subfigure}{0.24\linewidth}
        \centering
    \includegraphics[width=2.7cm,height=2cm]{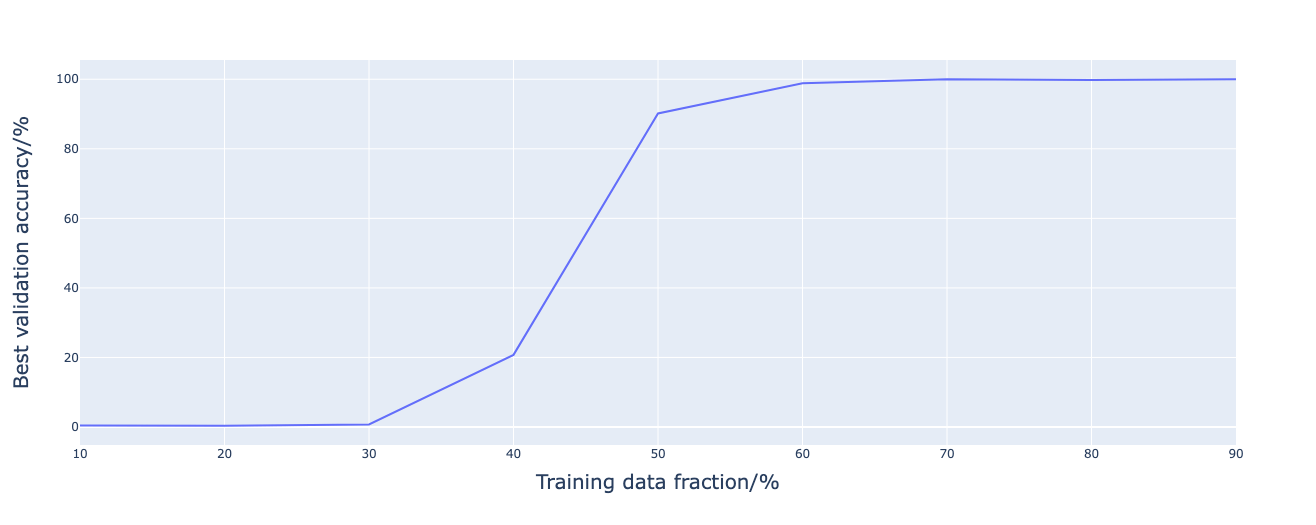}
        \caption{\scriptsize Minibatch Adam}
    \end{subfigure} 
       \begin{subfigure}{0.24\linewidth}
        \centering
       \includegraphics[width=2.7cm,height=2cm]{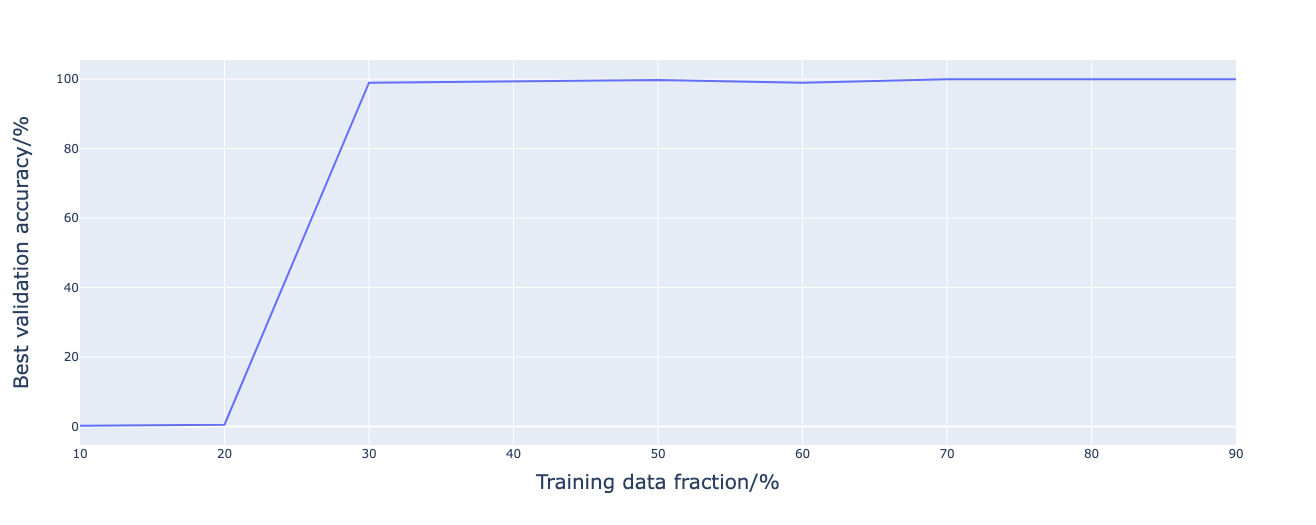}
        \caption{\scriptsize AdamW, weight decay 1}
    \end{subfigure}
    \begin{subfigure}{0.24\linewidth}
        \centering
\includegraphics[width=2.7cm,height=2cm]{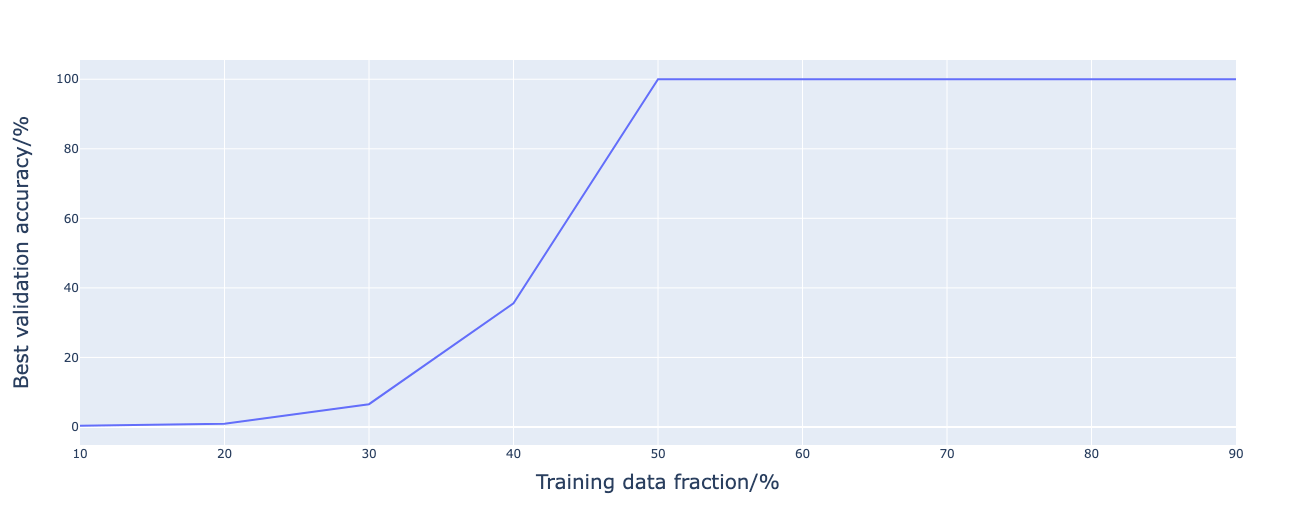}
        \caption{\scriptsize Full Batch AdamW}
    \end{subfigure}
   
    \caption{Different optimization  algorithms lead to different amount of generalization within an optimization budget of 1800 steps for the problem of (x+y) mod 97. Weight decay, i.e. AdamW, improves generalization the most, but some generalization happens even with full batch optimizers and models without weight decay or activation noise at high percentages of training data fraction. Suboptimal choice hyperparameters severely limit generalization. Note this: training accuracy achieved after approximately 100-200 updates for all optimization methods and training data fractions.}
    \label{fig:algopic}
\end{figure}

We observed different optimization algorithms to the problem and the result is displayed in Figure \ref{fig:algopic}. We used the simplified pretrained Transformer model stated in Section \ref{section:modelopt} with a total of about $10^5$ non-embedding parameters.

For final configuration of hyperparameters, we have chosen a balance of performance and we chose not to anneal the learning rate for the experiments in the previous paper even though it performed better in some situations. For most experiments, we used AdamW optimizers with learning rate $10^{-3}$, weight decay 1, $\beta_1=0.9, \beta_2=0.98$, linear learning rate warmup over the first 10 updats, full batch and optimization budget of 1800 gradient updates. 

In Figure \ref{fig:algopic}, we have tried the following variants(listed in the reading order):

\begin{itemize}
\item Adam optimizer with learning rate $3 \cdot 10^{-4}$
\item Adam optimizer with learning rate $10^{-3}$
\item Adam optimizer with learning rate $3\cdot 10^{-3}$
\item Adam optimizer with full batch and Gaussian noise added to the updated direction for each parameter ($W \gets W + \text{lr} \cdot (\Delta W+\epsilon)$, where $\epsilon $ is sampled from unit Gaussain, $\Delta W$ is the standard Adam weight update, and lr is the learning rate)
\item Adam optimizer on model with Gaussian weight noise of standard deviation 0.01 (each parameter $W$ replaced by $W+0.01 \cdot \epsilon$, with $\epsilon $ is sampled from unit Gaussain)
\item Adam optimizer with minibatch size 128
\item AdamW optimizer with weight decay 1
\item AdamW optimizer with minibatch size 128
\end{itemize}

Grokking is somewhat reminiscent of the phenomena of “epoch-wise”double descent \cite{power2022grokking},where generalization can improve after a period of over fitting. It has been found that regularization can mitigate double descent,similar perhaps to how weight decay influences grokking.

\section{Mechanisms of Grokking}
\label{section:grok}
Power et al. \cite{power2022grokking} conducted a study on neural networks, investigating their capacity to generalize to missing entries in binary operation tables. They introduced the concept of "grokking" and examined data efficiency curves for various binary operations. Significant research findings in this area have emerged, and we will now summarize two of these studies.

\paragraph{Structured representation and "Goldilocks zone"}:
In \cite{liu2022understanding}, it is suggested that structured representations play a pivotal role in the grokking phenomenon. This observation aligns with the emergence of structure in the learned input embeddings, as visualized in Figure 3 of \cite{power2022grokking}.

The study identifies four learning phases: comprehension, grokking, memorization, and confusion. The comprehension phase is characterized by the model's ability to learn the underlying structure of the data, while the grokking phase is characterized by delayed generalization due to the emergence of new structured representations. The memorization phase occurs when the model overfits the training data, and the confusion phase occurs when the model fails to learn any meaningful structure.

The study emphasizes the significance of the "Goldilocks zone," encompassing the comprehension and grokking phases, for achieving optimal representation learning and generalization performance. In the realm of representation learning, the "Goldilocks zone" denotes the range of hyperparameters that strike a balance between simplicity and complexity. Within this zone, the model can learn structured representations that are neither overly simplistic nor excessively complex. Consequently, this balanced state allows the model to achieve improved generalization performance by striking an equilibrium between model complexity and decoder simplicity.

Overall, the study highlights the importance of understanding the dynamics of structured representations during training and their impact on the phenomenon of grokking. It provides insights into how the "Goldilocks zone" impacts the learning phases and the emergence of structured representations, contributing to a deeper understanding of the dynamics of deep learning systems.

\paragraph{Implicit biases}: 
Lyu et al. (2023) investigated neural networks in classification and regression tasks \cite{lyu2023dichotomy}. They established that large initialization and small weight decay are essential conditions for the occurrence of the grokking phenomenon. However, specific examples using diagonal linear neural networks for classification tasks demonstrated the emergence of either grokking or \emph{misgrokking}.

Misgrokking implies that validation accuracy initially approaches near-perfection but experiences a sharp decline in the later stages of training. The degree of this match or mismatch significantly depends on the specific dataset, underscoring the importance of considering dataset characteristics in neural network analysis.

The researchers delved deeper into neural network behaviors, shedding light on the transition from optimal solutions to Karush-Kuhn-Tucker (KKT) solutions in misgrokking. In their exploration of this transition, they discovered that grokking can be influenced by the disparity between early-phase implicit bias and the alignment between late-phase implicit bias.

In summary, the study demonstrated how grokking or misgrokking can arise in classification and regression tasks based on the interaction between the network's implicit biases and the provided data.

\section{Conclusion}
\label{section:conclusion}
We explored the problem setting of grokking and conducted multifaceted experiments controlling various variables. In the experiments concerning the training data fraction, we observed that the task's difficulty decreased as the training data proportion increased, yet it still exhibited the same generalization patterns. In the experiments altering the model, we did not observe the grokking phenomenon. However, based on literature research, we understand this might be attributed to not employing weight decay techniques.

Besides, we’ve measured the accuracy for training different fractions of dataset with various forms of regularization. We present the data efficiency curves for a variety of interventions: full-batch gradient descent, stochastic gradient descent, large or small learning rates, weight decay \cite{DBLP:journals/corr/abs-1711-05101} and gradient noise \cite{neelakantan2015adding}. The results are presented in Figure \ref{fig:algopic}. 

While the phenomenon of grokking abnormal generalization was only discovered in 2022, there have been significant research findings in this area. Two of these mechanisms, namely Structured Representation and Implicit Bias, were discussed in this report.

Research on grokking mechanisms has the potential to analyze training dynamics and develop progress measures for predicting generalization. Furthermore, these findings provide insights into the optimization of neural networks and may have implications for a wide range of machine learning tasks.


\section*{Appendix}

\subsection*{Losses and Steps Until Generalization}
We provide additional details for the experiments conducted using a straightforward encoding method in conjunction with a Transformer model. The experiment is with mod 97.

The loss function is defined as follows:
\begin{align}\label{equ:loss}
    \mathcal{L}(p_{i,j},c_i)=-\frac{1}{N}\sum_{i=1}^{N}\log(p_{i,c_i}),
\end{align}
where $N$ represents the number of samples, $p_{i,j}$ is the predicted probability output by the model, and $c_i$ denotes the correct label indicated by the ground truth. 

\begin{figure}[H]
    \begin{subfigure}{0.32\linewidth}
        \centering
        \includegraphics[width=4cm,height=3cm]{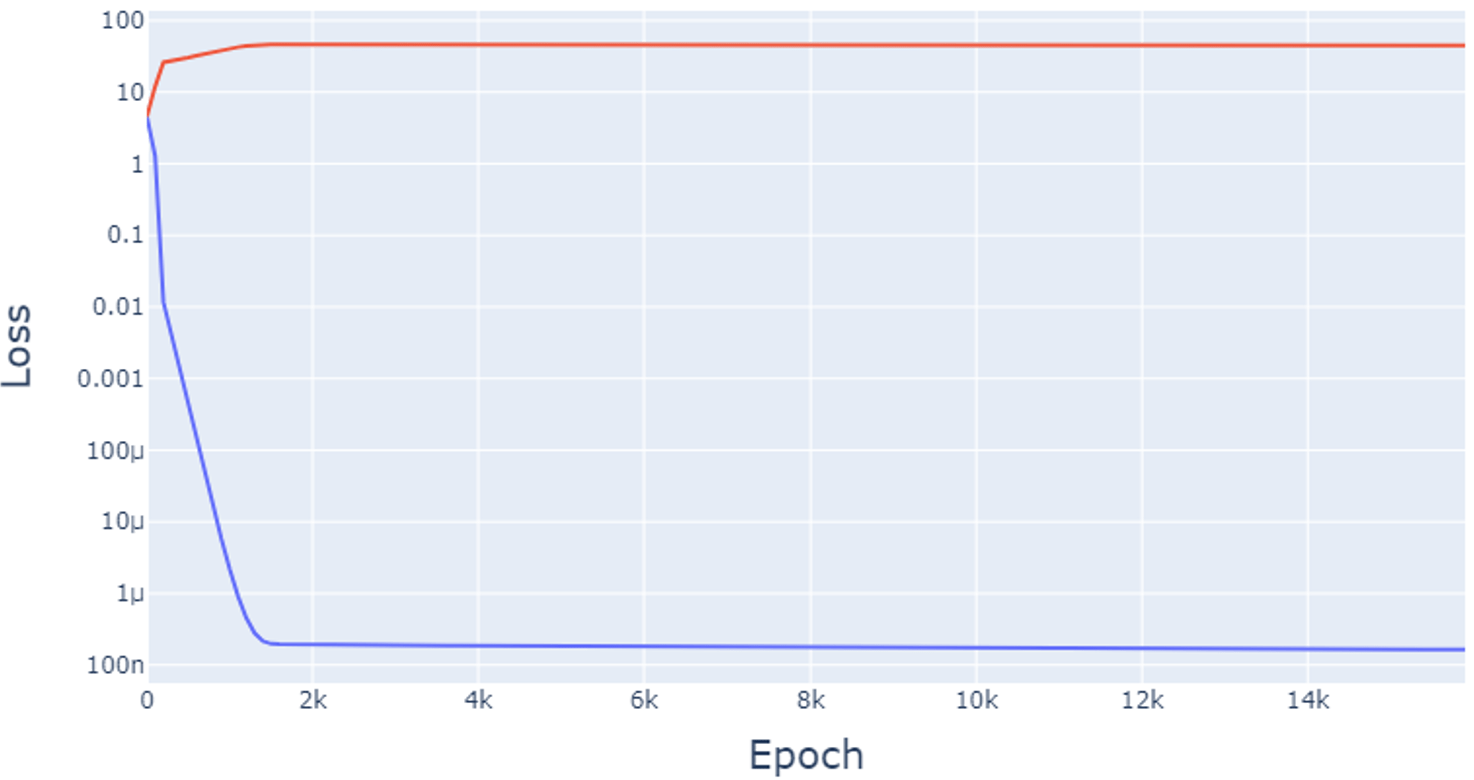}
        \caption{Loss at 15\% training data.}
    \end{subfigure}
    \begin{subfigure}{0.32\linewidth}
        \centering
        \includegraphics[width=4cm,height=3cm]{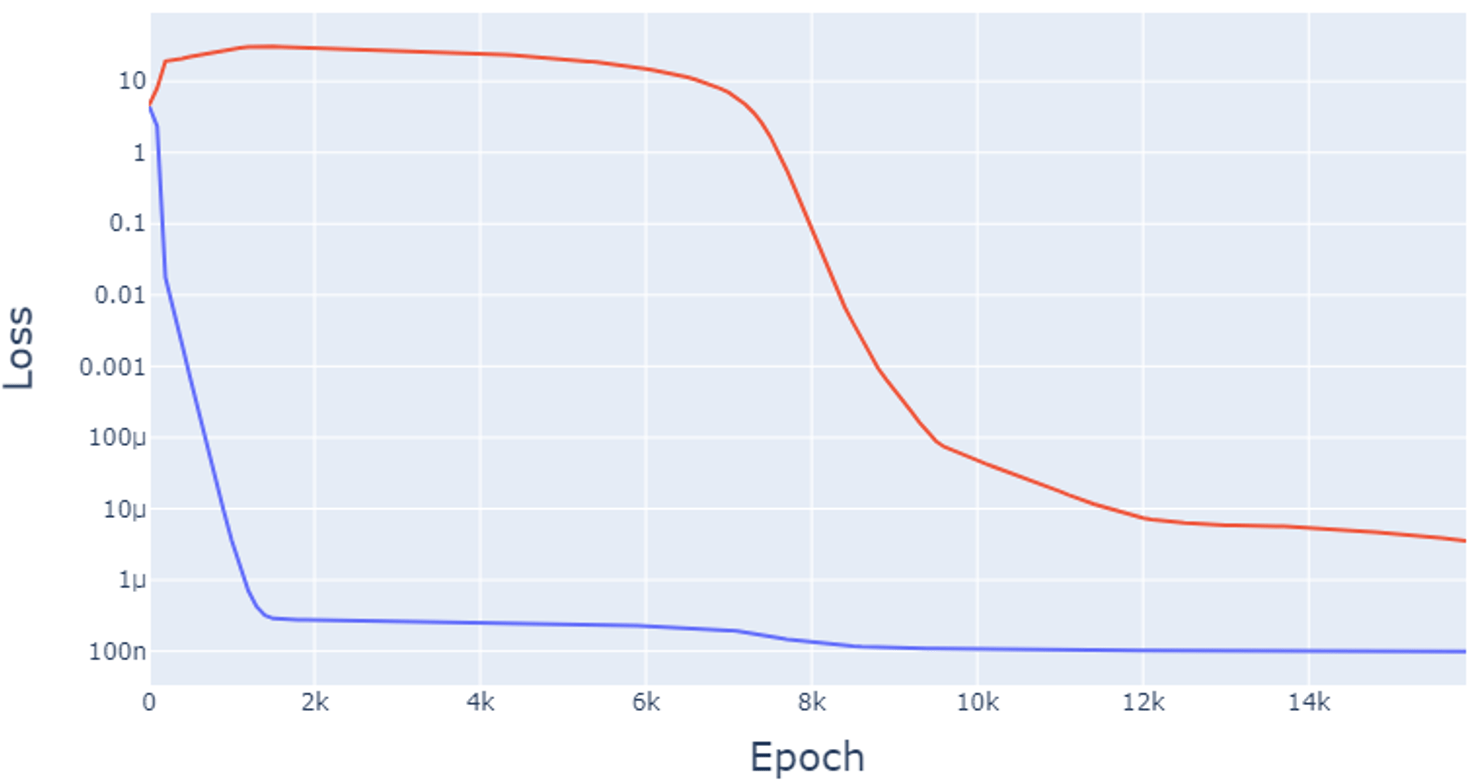}
        \caption{Loss at 30\% training data.}
    \end{subfigure}
    \begin{subfigure}{0.32\linewidth}
        \centering
        \includegraphics[width=4cm,height=3cm]{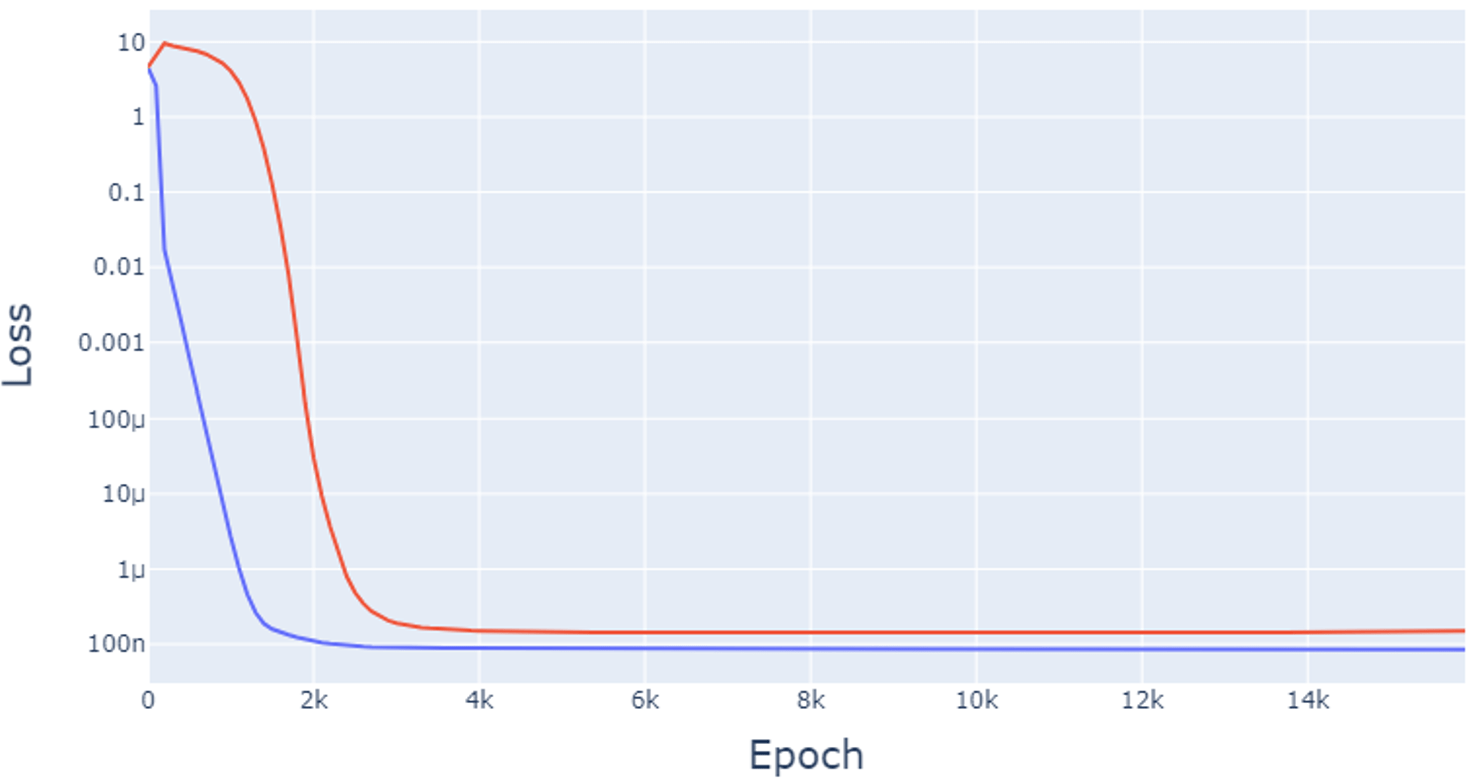}
        \caption{Loss at 45\% training data.}
    \end{subfigure}    
    \caption{Comparison of loss at different training data fractions. The blue lines for training and the red lines for validation.}
    \label{fig:loss_step_trans2}
\end{figure}
The changes in losses are synchronized with accuracy and also illustrate three typical generalization types.

\begin{figure}[H]
    \centering
    \includegraphics[width=8cm,height=5.5cm]{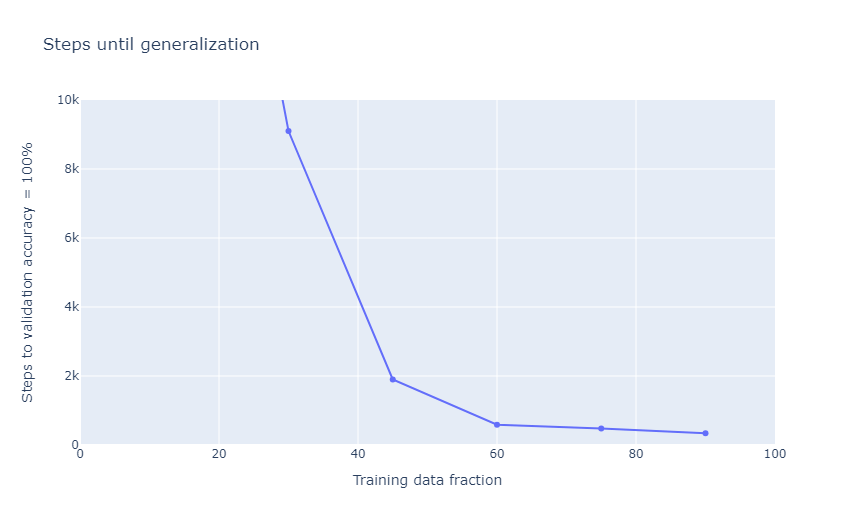}
    \caption{Steps until generalization at various training data fractions.}
    \label{fig:step_frac_trans2}
\end{figure}

\subsection*{Experiments with MLP Model}
The code for the MLP experiments in Subsection \ref{subsection:exp model} can be accessed here \url{https://colab.research.google.com/drive/1dF1hZlswdlwNfNm2VhigGDPeP5OiL1-Z?usp=sharing#scrollTo=6dAfNqolUN8D}.

For these experiments, we utilized the cross-entropy loss.

\begin{figure}[H]
    \begin{subfigure}{0.32\linewidth}
        \centering
        \includegraphics[width=4cm,height=3cm]{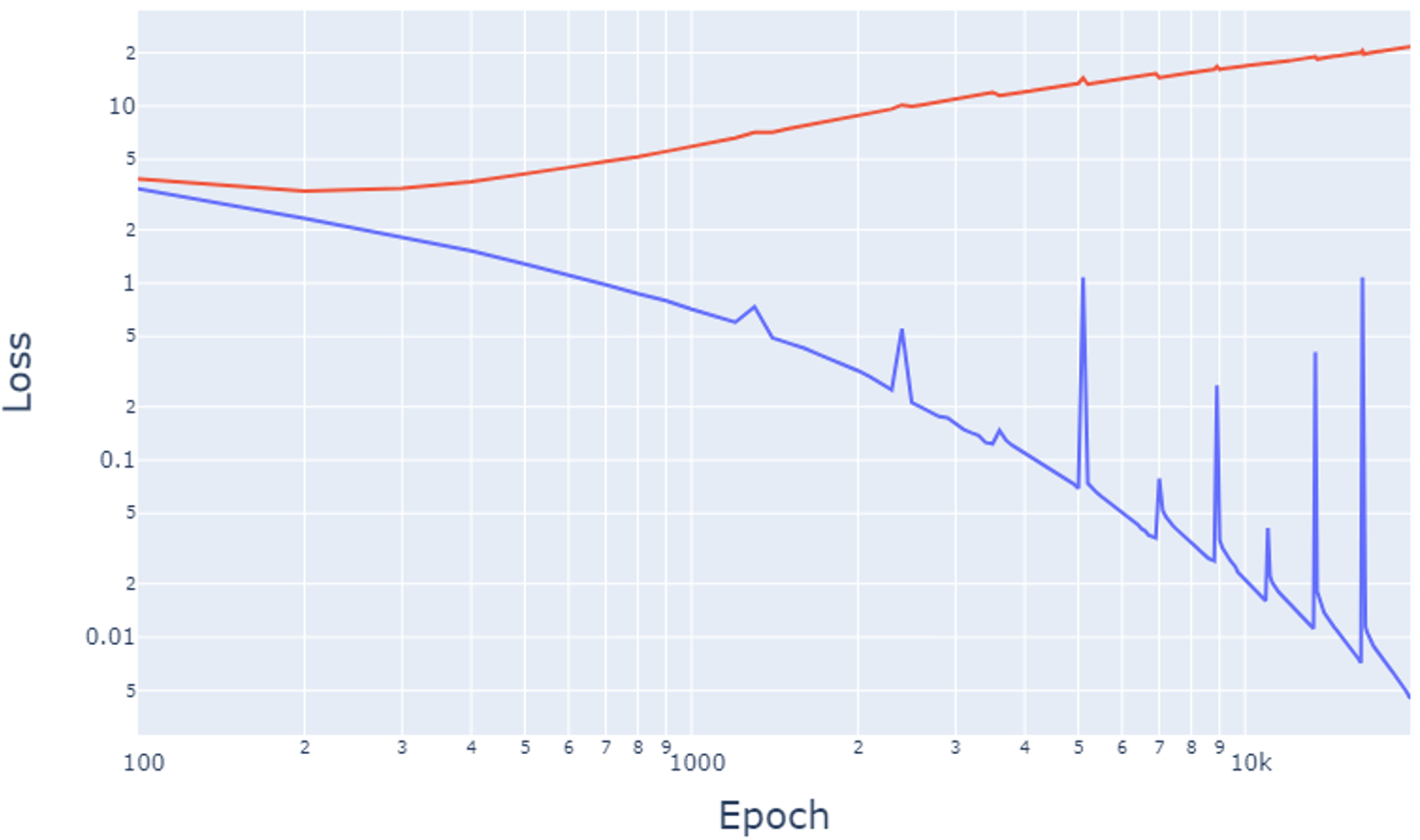}
        \caption{Loss at 15\% training data.}
    \end{subfigure}
    \begin{subfigure}{0.32\linewidth}
        \centering
        \includegraphics[width=4cm,height=3cm]{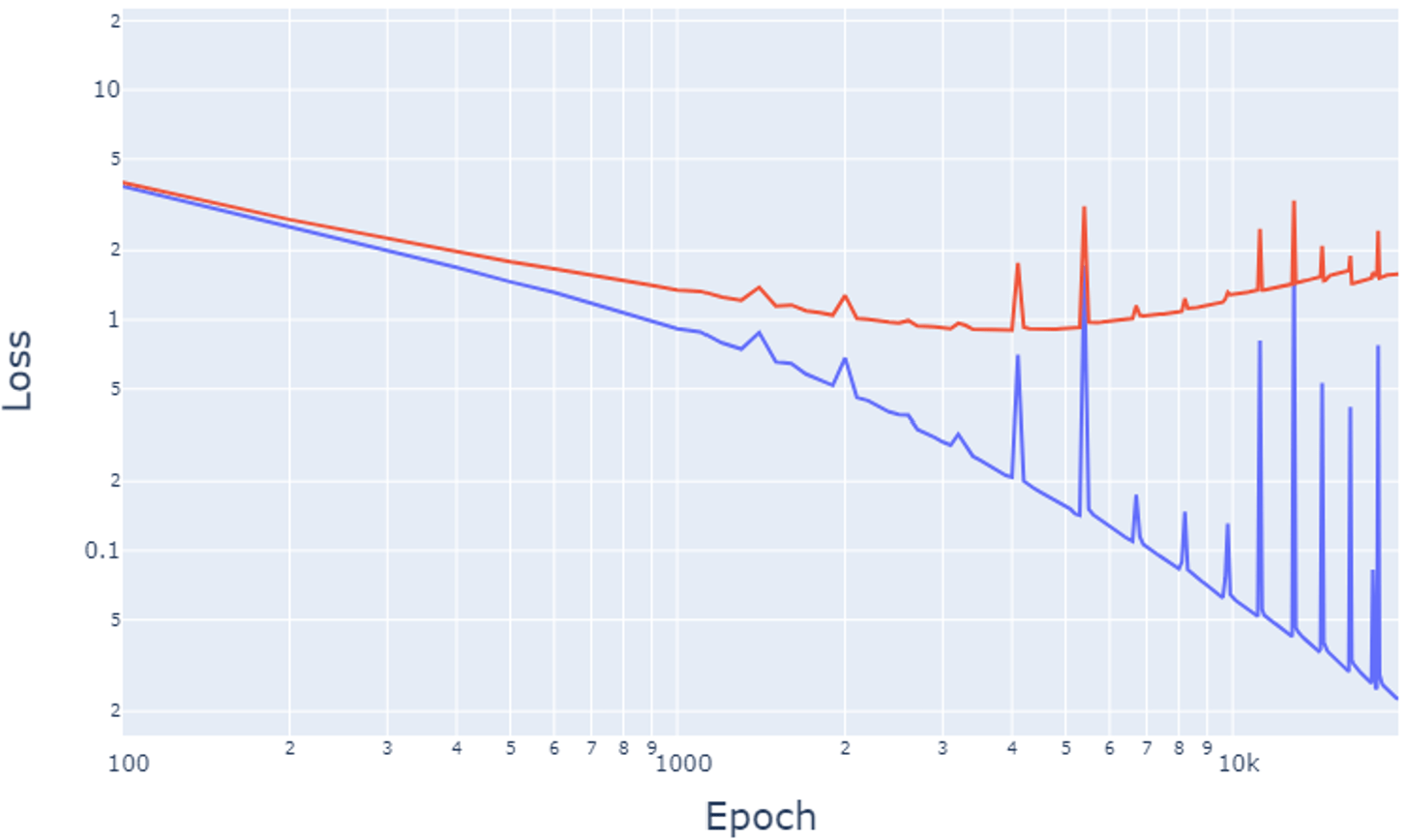}
        \caption{Loss at 30\% training data.}
    \end{subfigure}
    \begin{subfigure}{0.32\linewidth}
        \centering
        \includegraphics[width=4cm,height=3cm]{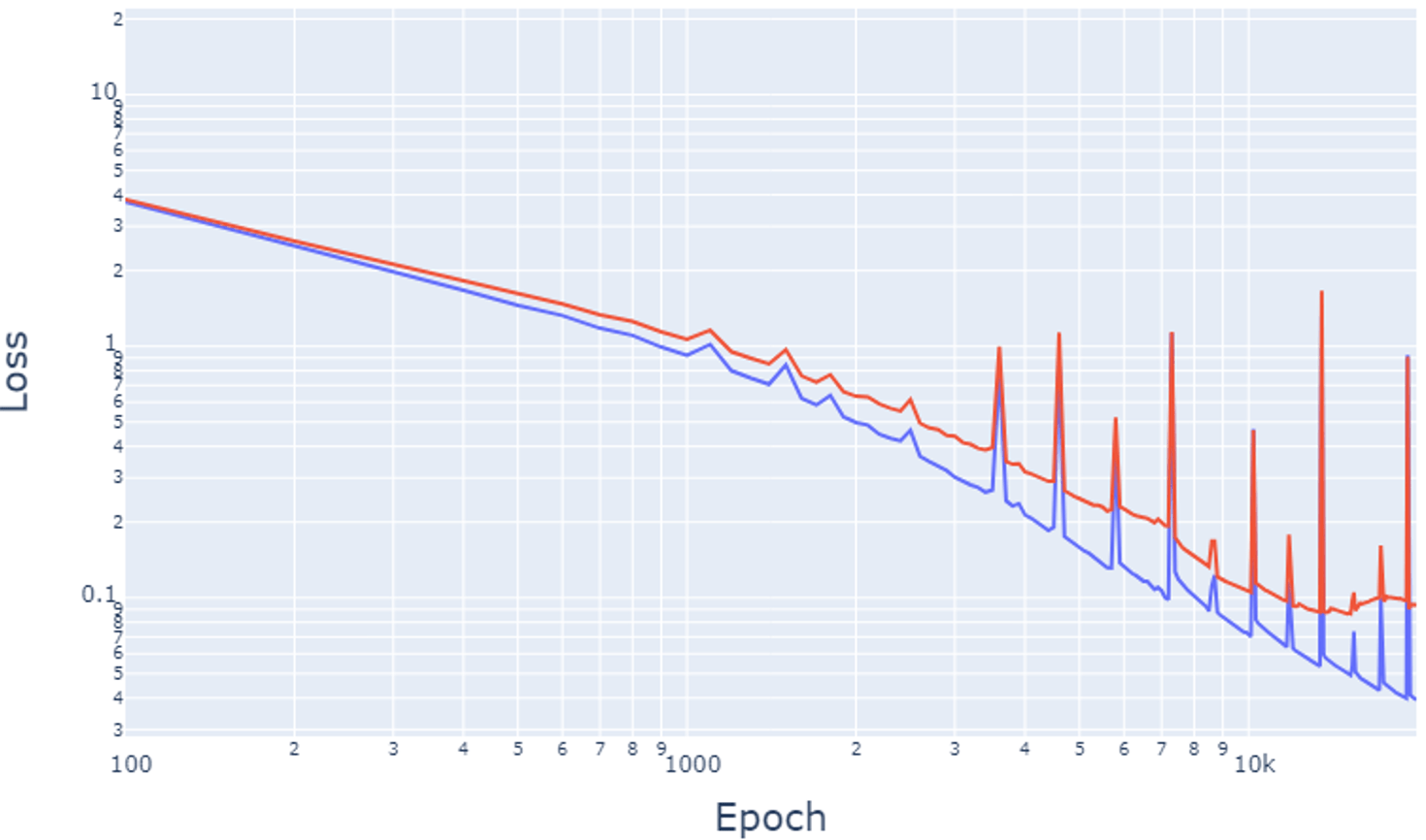}
        \caption{Loss at 45\% training data.}
    \end{subfigure}
    \caption{Comparison of loss at different training data fractions with MLP model. The blue lines for training and the red lines for validation.}
    \label{fig:loss_step_mlp}
\end{figure}
In the later stages of training, we observed an increase in the loss for the validation data.

\end{document}